\documentclass[11pt,a4paper]{article}
\usepackage[hyperref]{acl2019}
\usepackage{times}
\usepackage{latexsym}
\usepackage{linguex}

\usepackage{url}

\aclfinalcopy 

\usepackage{graphicx}
\usepackage{multirow}
\usepackage{amsmath,amsfonts,amssymb}
\DeclareMathOperator{\E}{\mathbb{E}}
\usepackage{bm}
\usepackage{array}
\usepackage{CJKutf8}
\usepackage{color}

\title{Sentence Centrality Revisited for Unsupervised Summarization}

\author{Hao Zheng \textnormal{and} Mirella Lapata\\
Institute for Language, Cognition and Computation\\
School of Informatics, University of Edinburgh\\
 10 Crichton Street, Edinburgh EH8 9AB\\
\texttt{Hao.Zheng@ed.ac.uk}~~~~\texttt{mlap@inf.ed.ac.uk}\\
}

\date{}

\makeatletter
\newcommand{\thickhline}{%
    \noalign {\ifnum 0=`}\fi \hrule height 1pt
    \futurelet \reserved@a \@xhline
}
\makeatother

\begin{document}
\maketitle
\begin{abstract}
  Single document summarization has enjoyed renewed interest in recent
  years thanks to the popularity of neural network models and the
  availability of large-scale datasets.  In this paper we develop an unsupervised approach arguing that it is unrealistic to expect
  large-scale and high-quality training data to be available or
  created for different types of summaries, domains, or languages.  We
  revisit a popular graph-based ranking algorithm and modify how node
  (aka sentence) centrality is computed in two ways: (a)~we employ
  BERT, a state-of-the-art neural representation learning model to
  better capture sentential meaning and (b)~we build graphs with
  directed edges arguing that the contribution of any two nodes to
  their respective centrality is influenced by their relative position
  in a document. Experimental results on three news summarization
  datasets representative of different languages and writing styles
  show that our approach outperforms strong baselines by a wide
  margin.\footnote{Our code is available at \url{https://github.com/mswellhao/PacSum}.}

\end{abstract}

\section{Introduction}

Single-document summarization is the task of generating a shorter
version of a document while retaining its most important content
\cite{nenkova2011automatic}. Modern neural network-based approaches
\cite{K16-1028, paulus2018a, AAAI1714636, P16-1046, P17-1099,
  N18-1158, D18-1443} have achieved promising results thanks to the
availability of large-scale datasets containing hundreds of thousands
of document-summary pairs
\cite{nytcorpus,hermann-nips15,newsroom-naacl18}.
Nevertheless, it is unrealistic to expect that large-scale and
high-quality training data will be available or created for different
summarization styles (e.g.,~highlights vs. single-sentence summaries),
domains (e.g., user- vs. professionally-written articles), and
languages.

It therefore comes as no surprise that unsupervised approaches have
been the subject of much previous research \cite{W97-0713, W00-0403, lin2002single, W04-3252, erkan2004lexrank, D08-1079, Wan2008MultidocumentSU, D13-1158, D15-1226, IJCAI1511225, AAAI1714613}. A very
popular algorithm for extractive single-document summarization is
TextRank \cite{W04-3252}; it represents document sentences as nodes in
a graph with \emph{undirected} edges whose weights are computed based
on sentence similarity. In order to decide which sentence to include
in the summary, a node's \emph{centrality} is often measured using
graph-based ranking algorithms such as PageRank \cite{ilprints361}.


In this paper, we argue that the centrality measure can be improved in
two important respects. Firstly, to better capture sentential meaning
and compute sentence similarity, we employ BERT \cite
{devlin2018bert}, a neural representation learning model which has
obtained state-of-the-art results on various natural language
processing tasks including textual inference, question answering, and
sentiment analysis. Secondly, we advocate that edges should be
\emph{directed}, since the contribution induced by two nodes'
connection to their respective centrality can be in many cases
unequal. For example, the two sentences below are semantically
related:

\ex. \label{ex:1} Half of hospitals are letting patients jump
 NHS queues for cataract surgery if they pay for it themselves, an investigation has revealed. 

\ex. \label{ex:2} Clara Eaglen, from the royal national institute of blind people,
  said: ``It's shameful that people are being asked to consider
  funding their own treatment when they are entitled to it for free,
  and in a timely manner on the NHS.''

  Sentence~\ref{ex:1} describes a news event while sentence~\ref{ex:2}
  comments on it. Sentence~\ref{ex:2} would not make much sense on its
  own, without the support of the preceding sentence, whose content is
  more central.  Similarity as an undirected measure, cannot
  distinguish this fundamental intuition which is also grounded in
  theories of discourse structure \cite{mann1988rhetorical}
  postulating that discourse units are characterized in terms of their
  text importance: \emph{nuclei} denote central segments, whereas
  \emph{satellites} denote peripheral ones.

  We propose a simple, yet effective approach for measuring directed
  centrality for single-document summarization, based on the assumption that the contribution of any two nodes' connection to their respective centrality is influenced by their \emph{relative} position. Position information has been frequently
  used in summarization, especially in the news domain, either as a
  baseline that creates a summary by selecting the first~$n$ sentences
  of the document \cite{Nenkova:2005} or as a feature in
  learning-based systems
  \cite{A97-1042,Schilder:Kondadadi:2008,C10-2106}.  We transform
  undirected edges between sentences into directed ones by
  differentially weighting them according to their
  \emph{orientation}. Given a pair of sentences in the same document, one is looking
  forward (to the sentences following it), and the other is looking
  backward (to the sentences preceding it). For some types of
  documents (e.g.,~news articles) one might further expect sentences
  occurring early on to be more central and therefore backward-looking
  edges to have larger weights.




  We evaluate the proposed approach on three single-document news
  summarization datasets representative of different languages,
  writing conventions (e.g., important information is concentrated in
  the beginning of the document or distributed more evenly throughout)
  and summary styles (e.g.,~verbose or more telegraphic). We
  experimentally show that position-augmented centrality significantly
  outperforms strong baselines (including TextRank;
  \citealt{W04-3252}) across the board.  In addition, our best system
  achieves performance comparable to supervised systems trained on
  hundreds of thousands of examples \cite{N18-1158,P17-1099}. We
  present an alternative to more data-hungry models, which we argue
  should be used as a standard comparison when assessing the merits of
  more sophisticated supervised approaches over and above the baseline
  of extracting the leading sentences (which our model outperforms).

  Taken together, our results indicate that directed centrality
  improves the selection of salient content
  substantially. Interestingly, its significance for unsupervised
  summarization has gone largely unnoticed in the research
  community. For example, \texttt{gensim}
  \cite{barrios2016variations}, a widely used open-source
  implementation of TextRank only supports building undirected graphs,
  even though follow-on work \cite{P04-3020} experiments with
  position-based directed graphs similar to ours. Moreover, our
  approach highlights the effectiveness of pretrained embeddings for
  the summarization task, and their promise for the development of
  unsupervised methods in the future. We are not aware of any previous
  neural-based approaches to unsupervised single-document
  summarization, although some effort has gone into developing
  unsupervised models for multi-document summarization using
  reconstruction objectives
  \cite{AAAI1714613,C16-1143,DBLP:journals/corr/abs-1810-05739}.




 %

\section{Centrality-based  Summarization}

\subsection{Undirected Text Graph}
A prominent class of approaches in unsupervised summarization uses
graph-based ranking algorithms to determine a sentence's salience for
inclusion in the summary \cite{W04-3252, erkan2004lexrank}. A document
(or a cluster of documents) is represented as a graph, in which nodes
correspond to sentences and edges between sentences are weighted by
their similarity. A node's centrality can be measured by simply
computing its degree or running a ranking algorithm such as PageRank
\cite{ilprints361}.

For single-document summarization, let $D$~denote a
document consisting of a sequence of sentences $\{s_1,s_2,...,s_n\}$,
and $e_{ij}$~the similarity score for each pair $(s_i,s_j)$. The
degree centrality for sentence $s_i$ can be defined as:
\begin{eqnarray}
 \textbf{centrality}(s_i) = \sum_{j \in \{1,..,i-1,i+1,..,n\}} e_{ij}
\end{eqnarray}
After obtaining the centrality score for each sentence, sentences are
sorted in reverse order and the top ranked ones are included in the
summary.

TextRank \cite{W04-3252} adopts PageRank \cite{ilprints361} to compute
node centrality recursively based on a Markov chain model. Whereas
degree centrality only takes local connectivity into account, PageRank
assigns relative scores to all nodes in the graph based on the
recursive principle that connections to nodes having a high score
contribute more to the score of the node in question. Compared to
degree centrality, PageRank can in theory be better since the global
graph structure is considered. However, we only observed marginal
differences in our experiments (see
Sections~\ref{sec:experimental-setup} and~\ref{sec:results} for
details).

\subsection{Directed Text Graph}

The idea that textual units vary in terms of their importance or
salience, has found support in various theories of discourse structure
including Rhetorical Structure Theory (RST;
\citealt{mann1988rhetorical}).  RST is a compositional model of
discourse structure, in which elementary discourse units are combined
into progressively larger discourse units, ultimately covering the
entire document.  Discourse units are linked to each other by
rhetorical relations (e.g., \emph{Contrast}, \emph{Elaboration}) and
are further characterized in terms of their text importance:
\emph{nuclei} denote central segments, whereas \emph{satellites}
denote peripheral ones. The notion of nuclearity has been leveraged
extensively in document summarization \cite{W97-0713, W98-1124,
  D13-1158} and in our case provides motivation for taking
directionality into account when measuring centrality.




We could determine nuclearity with the help of a discourse parser
(\citealt{li-li-chang:2016:EMNLP2016,feng-hirst:2014:P14-1,joty-EtAl:2013:ACL2013,liu-lapata:2017:EMNLP2017},
inter alia) but problematically such parsers rely on the availability
of annotated corpora as well as a wider range of standard NLP tools
which might not exist for different domains, languages, or text
genres.  We instead approximate nuclearity by relative position in the
hope that sentences occurring earlier in a document should be more
central. Given any two sentences $s_i,s_j$ $(i < j)$ taken from the
same document $D$, we formalize this simple intuition by transforming
the undirected edge weighted by the similarity score~$e_{ij}$
between~$s_i$ and~$s_j$ into two directed ones differentially weighted
by~$\lambda_1 e_{ij}$ and~$\lambda_2 e_{ij}$. Then, we can refine the
centrality score of~$s_i$ based on the directed graph as follows:
\begin{eqnarray}
\label{eq:dircentral}
 \textbf{centrality}(s_i) = \lambda_1 \sum_{j<i} e_{ij}   + \lambda_2 \sum_{j>i} e_{ij}
\end{eqnarray}
where $\lambda_1$, $\lambda_2$ are different weights for forward- and
backward-looking directed edges. Note that when $\lambda_1$ and
$\lambda_1$ are equal to~1, Equation~\eqref{eq:dircentral} becomes
degree centrality. The weights can be tuned experimentally on a
validation set consisting of a small number of documents and
corresponding summaries, or set manually to reflect prior knowledge
about how information flows in a document. During tuning experiments,
we set $\lambda_1 + \lambda_2 = 1$ to control the number of free
hyper-parameters. Interestingly, we find that the optimal $\lambda_1$
tends to be negative, implying that similarity with previous content
actually hurts centrality. This observation contrasts with existing
graph-based summarization approaches \cite{W04-3252,P04-3020} where
nodes typically have either no edge or edges with positive
weights. Although it is possible to use some extensions of PageRank
\cite{kerchove2008pagetrust} to take negative edges into account, we
leave this to future work and only consider the definition of
centrality from Equation~(\ref{eq:2}) in this paper.

\section{Sentence Similarity Computation}
\label{sec:sent-simil-comp}

The key question now is how to compute the similarity between two
sentences. There are many variations of the similarity function of
TextRank \cite{barrios2016variations} based on symbolic sentence
representations such as \mbox{tf-idf}. We instead employ a
state-of-the-art neural representation learning model.  We use BERT
\cite{devlin2018bert} as our sentence encoder and fine-tune it based
on a type of sentence-level distributional hypothesis
\cite{harris1954distributional, polajnar2015exploration} which we
explain below. Fine-tuned BERT representations are subsequently used
to compute the similarity between sentences in a document.

\subsection{BERT as Sentence Encoder}
\label{sec:bert-as-sentence}

We use BERT (Bidirectional Encoder Representations from Transformers;
\citealt{devlin2018bert}) to map sentences into deep continuous
representations. BERT adopts a multi-layer bidirectional Transformer
encoder \cite{vaswani2017attention} and uses two unsupervised
prediction tasks, i.e.,~masked language modeling and next sentence
prediction, to pre-train the encoder.

The language modeling task aims to predict masked tokens by jointly
conditioning on both left and right context, which allows pre-trained
representations to fuse both contexts in contrast to conventional
uni-directional language models. Sentence prediction aims to model the
relationship between two sentences. It is a binary classification task,
essentially predicting  whether the second sentence in a sentence pair is
indeed the next sentence. 
Pre-trained BERT representations can be fine-tuned with just one
additional output layer to create state-of-the-art models for a wide
range of tasks, such as question answering and language inference. We use BERT to encode sentences for unsupervised summarization.

\subsection{Sentence-level Distributional Hypothesis}
\label{sec:sent-level-distr}

To fine-tune the BERT encoder, we exploit a type of sentence-level
distributional hypothesis
\cite{harris1954distributional,polajnar2015exploration} as a means to
define a training objective. In contrast to skip-thought vectors
\cite{NIPS2015_5950} which are learned by reconstructing the
surrounding sentences of an encoded sentence, we borrow the idea of
negative sampling from word representation learning
\cite{NIPS2013_5021}.  Specifically, for a sentence~$s_i$ in  document~$D$, we take its previous sentence~$s_{i-1}$ and its
following sentence $s_{i+1}$ to be positive examples, and consider any other sentence in the corpus to be a negative example.  The
training objective for~$s_i$ is defined as:
\begin{eqnarray}
\label{eq:objective}
\log \sigma({v_{s_{i-1}}^{\prime}}^\top v_{s_i})  +  \log\sigma({v_{s_{i+1}}^{\prime}}^\top v_{s_i}) \nonumber \\
+  \E_{s \sim P(s)} \big [ \log \sigma({-v_s^{\prime}}^\top v_s) ] 
\end{eqnarray}
where $v_s$ and $v_s^{\prime}$ are two different representations of
sentence $s$ via two differently parameterized BERT encoders; $\sigma$
is the sigmoid function; and~$P(s)$ is a uniform distribution defined
over the sentence space.

The objective in Equation~\eqref{eq:objective} aims to distinguish
context sentences from other sentences in the corpus, and the encoder
is pushed to capture the meaning of the intended sentence in order to
achieve that. We sample five negative samples for each positive
example to approximate the expectation. Note, that this approach is
much more computationally efficient, compared to reconstructing
surrounding sentences \cite{NIPS2015_5950}.


\subsection{Similarity Matrix}


Once we obtain representations $\{v_1, v_2,...,v_n\}$ for sentences
$\{s_1 , s_2, \dots, s_n\}$ in document~$D$, we employ pair-wise dot
product to compute an unnormalized similarity
matrix~$\bar{\mathrm{E}}$:
\begin{eqnarray}
\bar{\mathrm{E}}_{ij} = {v_i}^\top v_j
\end{eqnarray}
We could also use cosine similarity, but we empirically found that the
dot product performs better.

The final normalized similarity matrix $\mathrm{E}$ is defined based on $\bar{\mathrm{E}}$:
\begin{eqnarray}
\tilde{\mathrm{E}}_{ij}    &=& \bar{\mathrm{E}}_{ij} - \Big[\min{\bar{\mathrm{E}}} + \beta (\max{\bar{\mathrm{E}}} - \min{\bar{\mathrm{E}}}) \Big]~ \quad  \label{eq:1}  \\
\mathrm{E}_{ij} &=& 
\begin{cases}
    \tilde{\mathrm{E}}_{ij}  & \quad \text{if } \tilde{\mathrm{E}}_{ij} > 0 \\
    0  & \quad  \text{otherwise}
 \end{cases}
 \label{eq:2}
\end{eqnarray}
Equation (\ref{eq:1}) aims to remove the effect of absolute values by
emphasizing the relative contribution of different similarity
scores. This is particularly important for the adopted sentence
representations which in some cases might assign very high values to
all possible sentence pairs. Hyper-parameter~$\beta$ $(\beta \in
[0,1])$ controls the threshold below which the similarity score is set
to~0.

\section{Experimental Setup}
\label{sec:experimental-setup}

In this section we present our experimental setup for evaluating our
unsupervised summarization approach which we call \textsc{PacSum} as a
shorthand for \textbf{P}osition-\textbf{A}ugmented \textbf{C}entrality
based \textbf{Sum}marization.

\subsection{Datasets} 

\begin{table}[t]

\resizebox{\linewidth}{!}{
\begin{tabular}{l|r|rc|cc}
\thickhline
\multirow{2}{*}{{\bf Dataset}}  & \multirow{2}{*}{{ \# docs}} &  \multicolumn{2}{c|}{{ avg. document}} & \multicolumn{2}{c}{{ avg. summary}} \\
                  &  &  words   &  sen. & words & sen.  \\
                  \thickhline
                  
                  CNN+DM &  11,490  & 641.9   & 28.0 & 54.6   & 3.9  \\
                  NYT &     4,375   & 1,290.5 & 50.7 & 79.8   & 3.5 \\
                  TTNews &  2,000   & 1,037.1 & 21.8 & 44.8   & 1.1 \\
                  
  \thickhline
  
\end{tabular}
}
\caption{Statistics on  NYT,  CNN/Daily Mail, and TTNews datasets
  (test set). We compute the average document and
  summary length in terms of number of words and sentences, respectively.} \label{table:dataset_stats}
  \vspace{-0.1in}
\end{table}


We performed experiments on three recently released single-document
summarization datasets representing different languages, document
information distribution, and summary
styles. Table~\ref{table:dataset_stats} presents statistics on these
datasets (test set); example summaries are shown in
Table~\ref{table:example}.

The CNN/DailyMail dataset \cite{NIPS2015_5945} contains news articles
and associated highlights, i.e., a few bullet points giving a brief
overview of the article. We followed the standard splits for training,
validation, and testing used by supervised systems (90,266/1,220/1,093
CNN documents and 196,961/12,148/10,397 DailyMail documents). We did
not anonymize entities.

The \textsc{Lead-3} baseline (selecting the first three sentences in
each document as the summary) is extremely difficult to beat on
CNN/DailyMail \cite{N18-1158,D18-1206}, which implies that salient
information is mostly concentrated in the beginning of a document. NYT
writers follow less prescriptive
guidelines\footnote{\url{https://archive.nytimes.com/www.nytimes.com/learning/issues_in_depth/10WritingSkillsIdeas.html}},
and as a result salient information is distributed more evenly in the
course of an article \cite{P16-1188}. We therefore view the NYT
annotated corpus \cite{nytcorpus} as complementary to CNN/DailyMail in
terms of evaluating the model's ability of finding salient
information. We adopted the training, validation and test splits
(589,284/32,736/32,739) widely used for evaluating abstractive
summarization systems. However, as noted in \citet{P16-1188}, some
summaries are extremely short and formulaic (especially those for
obituaries and editorials), and thus not suitable for evaluating
extractive summarization systems. Following \citet{P16-1188}, we
eliminate documents with summaries shorter than~50 words. As a result,
the NYT test set contains longer and more elaborate summary sentences
than CNN/Daily Mail (see Table~\ref{table:dataset_stats}).

Finally, to showcase the applicability of our approach across
languages, we also evaluated our model on TTNews
\cite{hua2017overview}, a Chinese news summarization corpus, created
for the shared summarization task at NLPCC 2017. The corpus contains a
large set of news articles and corresponding human-written summaries
which were displayed on the Toutiao app (a mobile news app). Because
of the limited display space on the mobile phone screen, the summaries
are very concise and typically contain just one sentence. There are
50,000 news articles with summaries and 50,000 news articles without
summaries in the training set, and 2,000 news articles in test set.

\subsection{Implementation Details}
\label{sec:impl-deta}

For each dataset, we used the documents in the training set to
fine-tune the BERT model; hyper-parameters ($\lambda_1, \lambda_2,
\beta$) were tuned on a validation set consisting of 1,000 examples
with gold summaries, and model performance was evaluated on the test
set.

We used the publicly released BERT model\footnote{\url{https://github.com/google-research/bert}} \cite{devlin2018bert} to initialize our sentence
encoder. English and Chinese versions of BERT were respectively used
for the English and Chinese corpora. As mentioned in
Section~\ref{sec:sent-level-distr}, we fine-tune BERT using negative
sampling; we randomly sample five negative examples for every positive
one to create a training instance.  Each mini-batch included 20~such
instances, namely 120 examples. We used Adam \cite{kingma2014adam} as
our optimizer with initial learning rate set to~4e-6.

\section{Results}
\label{sec:results}

\subsection{Automatic Evaluation}

We evaluated summarization quality automatically using ROUGE F1
\cite{rouge}. We report unigram and bigram overlap (\mbox{ROUGE-1} and
\mbox{ROUGE-2}) as a means of assessing informativeness and the
longest common subsequence (\mbox{ROUGE-L}) as a means of assessing
fluency.

\paragraph{NYT and CNN/Daily Mail}

\begin{table*}[t]
\centering
\begin{tabular}{l|ccc|ccc}
\thickhline
\multirow{2}{*}{{\bf Method}}  & \multicolumn{3}{c|}{{\bf NYT}} & \multicolumn{3}{c}{{\bf CNN+DM}} \\
                  &   R-1 &  R-2 & R-L & R-1 & R-2 & R-L  \\
                  \thickhline
                  \textsc{Oracle}                  & 61.9 & 41.7 & 58.3     & 54.7 & 30.4 & 50.8 \\
 
                  \textsc{Refresh}\footnotemark{} \cite{N18-1158}   & 41.3 & 22.0 & 37.8      & 41.3 & 18.4 & 37.5  \\
                  \textsc{Pointer-Generator} \cite{P17-1099}      & 42.7 & 22.1 & 38.0      & 39.5 & 17.3 & 36.4  \\ \hline
                  \textsc{Lead-3}                  & 35.5 & 17.2 & 32.0     & 40.5 & 17.7 & 36.7  \\
                  \textsc{Degree} (\mbox{tf-idf})       & 33.2  & 13.1 &  29.0    & 33.0 & 11.7 & 29.5 \\
                  \textsc{TextRank} (\mbox{tf-idf})                & 33.2 & 13.1 & 29.0     & 33.2 & 11.8 & 29.6  \\
                  \textsc{TextRank} (skip-thought vectors)   & 30.1 & 9.6 & 26.1     & 31.4 & 10.2 & 28.2  \\
                  \textsc{TextRank} (BERT)                & 29.7 & 9.0 & 25.3     & 30.8 & 9.6 & 27.4  \\ \hline
 
\textsc{PacSum} (\mbox{tf-idf})    & 40.4 & 20.6 & 36.4     & 39.2 & 16.3 & 35.3 \\
\textsc{PacSum} (skip-thought vectors) & 38.3 & 18.8 & 34.5     & 38.6 & 16.1 &34.9 \\
\textsc{PacSum} (BERT)            & 41.4 & 21.7 & 37.5     & 40.7
                                 &17.8&36.9 \\ \thickhline
  
\end{tabular}
\caption{Test set results on the NYT and CNN\/DailyMail datasets using
  ROUGE F1 (R-1 and R-2 are shorthands for unigram and bigram overlap,
  R-L is the longest common subsequence).} \label{table:nc_results} 
\end{table*}

Table~\ref{table:nc_results} summarizes our results on the NYT and
CNN/Daily Mail corpora (examples of system output can be found in the
Appendix).  We forced all extractive approaches to select three
summary sentences for fair comparison. The first block in the table
includes two state-of-the-art supervised models. \textsc{Refresh}
\cite{N18-1158} is an extractive summarization system trained by
globally optimizing the ROUGE metric with reinforcement
learning. \textsc{Pointer-Generator} \cite{P17-1099} is an abstractive
summarization system which can copy words from the source text while
retaining the ability to produce novel words. As an upper bound, we
also present results with an extractive oracle system. We used a
greedy algorithm similar to \citet{AAAI1714636} to generate an oracle
summary for each document. The algorithm explores different
combinations of sentences and generates an oracle consisting of
multiple sentences which maximize the ROUGE score against the gold
summary.

The second block in Table~\ref{table:nc_results} presents the results
of the \textsc{Lead-3} baseline (which simply creates a summary by
selecting the first three sentences in a document) as well as various
instantiations of \textsc{TextRank} \cite{W04-3252}. Specifically, we
experimented with three sentence representations to compute sentence
similarity. The first one is based on \mbox{tf-idf} where the value of
the corresponding dimension in the vector representation is the number
of occurrences of the word in the sentence times the idf (inverse
document frequency) of the word. Following \texttt{gensim}, We
pre-processed sentences by removing function words and stemming
words. The second one is based on the skip-thought model
\cite{NIPS2015_5950} which exploits a type of sentence-level
distributional hypothesis to train an encoder-decoder model trying to
reconstruct the surrounding sentences of an encoded
sentence.\footnotetext{The ROUGE scores here on CNN/Daily Mail are
  higher than those reported in the original paper, because we extract
  3 sentences in Daily Mail rather than 4.} We used the
publicly released skip-thought
model\footnote{\url{https://github.com/ryankiros/skip-thoughts}} to
obtain vector representations for our task. The third one is based on
BERT \cite{devlin2018bert} fine-tuned with the method proposed in this
paper. Finally, to determine whether the performance of PageRank and
degree centrality varies in practice, we also include a graph-based
summarizer with \textsc{Degree} centrality and \mbox{tf-idf}
representations.

The third block in Table~\ref{table:nc_results} reports results with
three variants of our model, \textsc{PacSum}. These include sentence
representations based on \mbox{tf-idf}, skip-thought vectors, and
BERT. Recall that \textsc{PacSum} uses directed degree centrality to
decide which sentence to include in the summary.  On both NYT and
CNN/Daily Mail datasets, \textsc{PacSum} (with BERT representations)
achieves the highest ROUGE F1 score, compared to other unsupervised
approaches. This gain is more pronounced on NYT where the gap between
our best system and \textsc{LEAD-3} is approximately 6~absolute
ROUGE-1 F1 points. Interestingly, despite limited access to only 1,000
examples for hyper-parameter tuning, our best system is comparable to
supervised systems trained on hundreds of thousands of examples (see
rows \textsc{Refresh} and \textsc{Pointer-Generator} in the table).

\begin{figure}[t]
\centering
\vspace{-0.1in}
\includegraphics[width=2.8in]{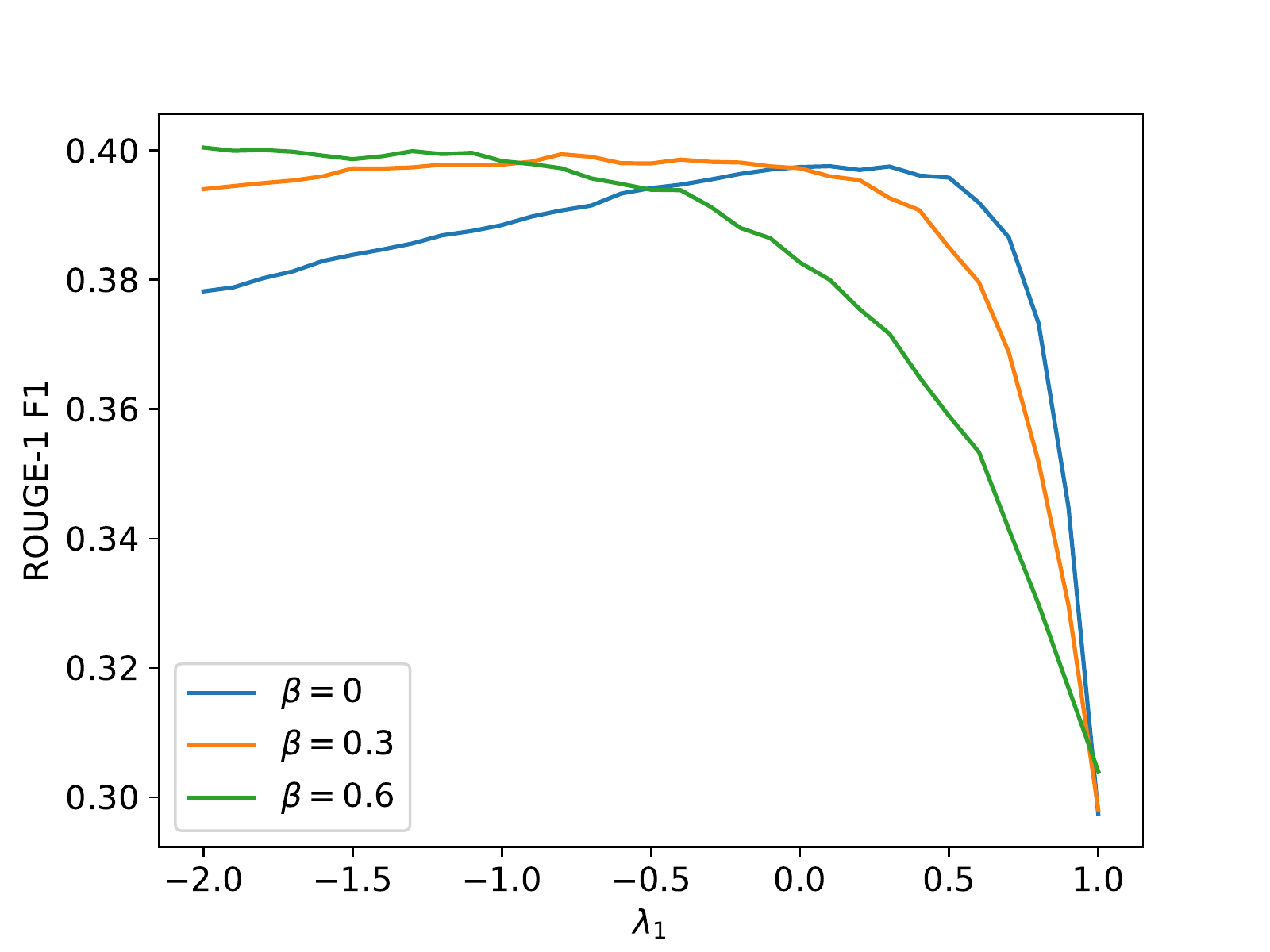}
\vspace{-0.1in}
\caption{\textsc{PacSum}'s performance against different values of
  $\lambda_1$ on the NYT validation set with with $\lambda_2 =
  1$.  Optimal hyper-parameters $(\lambda_1, \lambda_2, \beta)$  are $(-2, 1, 0.6)$.}
\label{figure_beta}
\vspace{-0.2in}
\end{figure}

As can be seen in Table~\ref{table:nc_results}, \textsc{Degree}
(\mbox{tf-idf}) is very close to \textsc{TextRank}
(\mbox{tf-idf}). Due to space limitations, we only show comparisons
between \textsc{Degree} and \textsc{TextRank} with \mbox{tf-idf},
however, we observed similar trends across sentence
representations. These results indicate that considering global
structure does not make a difference when selecting salient sentences
for NYT and CNN/Daily Mail, possibly due to the fact that news
articles in these datasets are relatively short (see Table
\ref{table:dataset_stats}).  The results in
Table~\ref{table:nc_results} further show that \textsc{PacSum}
substantially outperforms \textsc{TextRank} across sentence
representations, directly confirming our assumption that position
information is beneficial for determining sentence centrality in news
single-document summarization.  In Figure~\ref{figure_beta} we further
show how \textsc{PacSum}'s performance (ROUGE-1 F1) on the NYT
validation set varies as $\lambda_1$ ranges from -2 to 1
($\lambda_2 = 1$ and $\beta = 0, 0.3, 0.6$). The plot highlights that
differentially weighting a connection's contribution (via relative position)
has a huge impact on performance (ROUGE ranges from 0.30 to 0.40).  In
addition, the optimal $\lambda_1$ is negative, suggesting that
similarity with the previous content actually hurts centrality in this
case.


We also observed that \textsc{PacSum} improves further when equipped
with the BERT encoder. This validates the superiority of BERT-based
sentence representations (over \mbox{tf-idf} and skip-thought vectors)
in capturing sentence similarity for unsupervised
summarization. Interestingly, \textsc{TextRank} performs worse with
BERT. We believe that this is caused by the problematic centrality
definition, which fails to fully exploit the potential of continuous
representations. Overall, \textsc{PacSum} obtains improvements over
baselines on both datasets highlighting the effectiveness of our
approach across writing styles (highlights vs. summaries) and
narrative conventions. For instance, CNN/Daily Mail articles often
follow the inverted pyramid format starting with the most important
information while NYT articles are less prescriptive attempting to
pull the reader in with an engaging introduction and develop from
there to explain a topic.

\begin{table}[t]
\begin{tabular}{l|ccc}
  \thickhline
  \multirow{2}{*}{{\bf Method}}  & \multicolumn{3}{c}{{\bf TTNews}} \\
  &   R-1 &  R-2 & R-L   \\
  \thickhline                   
  \textsc{Oracle}                  & 45.6 & 31.4 & 41.7   \\
  \textsc{Pointer-Generator}     & 42.7 & 27.5 & 36.2   \\
  \textsc{Lead}                    & 30.8 & 18.4 & 24.9   \\
  \textsc{TextRank} (\mbox{tf-idf})        & 25.6 & 13.1 & 19.7   \\
  \textsc{PacSum} (BERT)  & 32.8 & 18.9 & 26.1 \\ \thickhline
  
\end{tabular}
\caption{Results on  Chinese  TTNews corpus using ROUGE F1 (R-1 and
  R-2 are shorthands for unigram and bigram overlap, R-L is the
  longest common subsequence).} \label{table:tt_results} 
\end{table}

\paragraph{TTNews Dataset}

Table~\ref{table:tt_results} presents our results on the TTNews corpus
using ROUGE F1 as our evaluation metric. We report results with
variants of \textsc{TextRank} (\mbox{tf-idf}) and \textsc{PacSum}
(BERT) which performed best on NYT and CNN/Daily Mail.  Since
summaries in the TTNews corpus are typically one sentence long (see
Table \ref{table:dataset_stats}), we also limit our extractive systems
to selecting a single sentence from the document. The \textsc{Lead}
baseline also extracts the first document sentence, while the
\textsc{Oracle} selects the sentence with maximum ROUGE score against
the gold summary in each document.  We use the popular
\textsc{Pointer-Generator} system of \citet{P17-1099} as a comparison
against supervised methods.

The results in Table~\ref{table:tt_results} show that
\textsc{Pointer-Generator} is superior to unsupervised methods, and
even comes close to the extractive oracle, which indicates that TTNews
summaries are more abstractive compared to the English
corpora. Nevertheless, even in this setting which disadvantages
extractive methods, \textsc{PacSum} outperforms \textsc{Lead} and
\textsc{TextRank} showing that our approach is generally portable
across different languages and summary styles. Finally, we show some examples of system output for the three datasets in Appendix.

\subsection{Human Evaluation}
\label{sec:human-evaluation}

\begin{table}[t]
\begin{tabular}{l|ccc}
  \thickhline
{\bf Method} &  {\bf NYT} & {\bf CNN+DM} & {\bf TTNews} \\
  \thickhline                   
  \textsc{Oracle}   & 49.0$^{*}$ & 53.9$^{*}$ & 60.0$^{*}$   \\
  \textsc{Refresh}  & \hspace*{-1ex}42.5 & \hspace*{-1ex}34.2 & ---   \\
  \textsc{Lead}     & 34.7$^{*}$ & 26.0$^{*}$ & 50.0$^{*}$   \\
  \textsc{PacSum}   & \hspace*{-1ex}44.4 & \hspace*{-1ex}31.1 & \hspace*{-1ex}56.0 \\ \thickhline
  
\end{tabular}
\caption{Results of QA-based evaluation on NYT, CNN/Daily Mail, and
  TTNews. We compute a system's final score as the average of all
  question scores.  Systems statistically significant from \textsc{PacSum}
  are denoted with an asterisk~* (using a one-way ANOVA
  with posthoc Tukey HSD tests; $p < 0.01$).} \label{table:human_results}
\end{table}

\begin{table*}[t]
\resizebox{\linewidth}{!}{
\begin{small}
\begin{tabular}{p{\textwidth}}

\thickhline
 \multicolumn{1}{c}{ \textsc{NYT}} \\\thickhline
  \textbf{Gold Summary:}  Marine Corps says that \textcolor{red}{V-22 Osprey}, hybrid aircraft with troubled past, will be sent to Iraq in September, where it will see combat for first time.
The Pentagon has placed so many restrictions on how it can be used in combat that plane -- which is \textcolor{red}{able to drop troops into battle like helicopter and then speed away like airplane} -- could have difficulty fulfilling marines longstanding mission for it. limitations on v-22, which cost \textcolor{red}{\$80 million apiece}, mean it can not evade enemy fire with same maneuvers and sharp turns used by helicopter pilots. \\
  \textbf{Questions:}  \hspace*{2ex}\textbullet\hspace{1ex} Which aircraft will be sent to Iraq?	\textcolor{red}{V-22 Osprey}  \\
    \hspace*{1.84cm}\textbullet\hspace*{1.4ex}What are the distinctive features of this type of aircraft?	\textcolor{red}{able to drop troops into battle like helicopter and then \hspace*{2.2cm}speed away like airplane}\\
\hspace*{1.84cm}\textbullet\hspace*{1.4ex}How much does each v-22 cost?	\textcolor{red}{\$80 million apiece}    \\\thickhline
   \multicolumn{1}{c}{}\\
\thickhline
\multicolumn{1}{c}{\textsc{CNN+DM}} \\ \thickhline
 \textbf{Gold Summary:} ``We're all equal, and we all deserve the same fair trial,'' says one juror.
The months-long murder trial of \textcolor{red}{Aaron Hernandez} brought jurors together.
\textcolor{red}{Foreperson}: ``It's been an incredibly emotional toll on all of us.'' \\
\textbf{Questions:}  \hspace*{2ex}\textbullet\hspace{1ex} Who was on trial?	\textcolor{red}{Aaron Hernandez} \\
 \hspace*{1.84cm}\textbullet\hspace*{1.4ex}Who said: ``It's been an incredibly emotional toll on all of us''? \textcolor{red}{Foreperson}
 \\\thickhline
   \multicolumn{1}{c}{}\\\thickhline
 \multicolumn{1}{c}{\textsc{TTNews}} \\\thickhline
  \textbf{Gold Summary :}  \begin{CJK*}{UTF8}{gbsn} 皇马今夏清洗名单曝光，\textcolor{red}{三}小将租借外出，科恩特朗、伊利亚拉门迪将被永久送出伯纳乌球场.
 (Real Madrid's cleaning list was exposed this summer, and the \textcolor{red}{three} players will be rented out. Coentrao and Illarramendi will permanently leave the Bernabeu Stadium.) \end{CJK*} \\
 \textbf{Question:} 
 \begin{CJK*}{UTF8}{gbsn}  皇马今夏清洗名单中几人将被外租？ \textcolor{red}{三} (How many people will be rented out by Real Madrid this summer? \textcolor{red}{three}) \end{CJK*} 
 \\\thickhline 
\end{tabular}

\end{small}
}
\caption{
NYT, CNN/Daily Mail and TTNews
  with corresponding questions. Words highlighted in red are answers
  to those questions. } \label{table:example} 

\vspace{-0.2in}
\end{table*}

In addition to automatic evaluation using ROUGE, we also evaluated
system output by eliciting human judgments. Specifically, we assessed
the degree to which our model retains key information from the
document following a question-answering (QA) paradigm which has been
previously used to evaluate summary quality and document compression
\cite{Clarke:Lapata:2010, N18-1158}. We created a set of questions
based on the gold summary under the assumption that it highlights the
most important document content. We then examined whether participants
were able to answer these questions by reading system summaries alone
without access to the article. The more questions a system can answer,
the better it is at summarizing the document.

For CNN/Daily Mail, we worked on the same~20 documents and associated
71~questions used in \citet{N18-1158}. For NYT, we randomly
selected~18 documents from the test set and created 59~questions in
total. For TTNews, we randomly selected 50~documents from the test set
and created 50~questions in total. Example questions (and answers) are
shown in Table~\ref{table:example}.  

We compared our best system \textsc{PacSum} (BERT) against
\textsc{Refresh}, \textsc{Lead-3}, and \textsc{Oracle} on CNN/Daily
Mail and NYT, and against \textsc{Lead-3} and \textsc{Oracle} on
TTNews. Note that we did not include \textsc{TextRank} in this
evaluation as it performed worse than \textsc{Lead-3} in previous
experiments (see Tables~\ref{table:nc_results}
and~\ref{table:tt_results}).  Five participants answered questions for
each summary. We used the same scoring mechanism from
\citet{N18-1158}, i.e., a correct answer was marked with a score of
one, partially correct answers with a score of~0.5, and zero
otherwise. The final score for a system is the average of all its
question scores. Answers for English examples were elicited using
Amazon’s Mechanical Turk crowdsourcing platform while answers for
Chinese summaries were assessed by in-house native speakers of
Chinese. We uploaded the data in batches (one system at a time) on AMT
to ensure that the same participant does not evaluate summaries from
different systems on the same set of questions.

The results of our QA evaluation are shown in
Table~\ref{table:human_results}. \textsc{Oracle}'s performance is
below~100, indicating that extracting sentences by maximizing ROUGE
fails in many cases to select salient content, capturing surface
similarity instead. \textsc{PacSum} significantly outperforms
\textsc{Lead} but is worse than \textsc{Oracle} which suggests there
is room for further improvement. Interestingly, \textsc{PacSum}
performs on par with \textsc{Refresh} (the two systems are not
significantly different).

\section{Conclusions}
\label{sec:conclusions}

In this paper, we developed an unsupervised summarization system which
has very modest data requirements and is portable across different
types of summaries, domains, or languages. We revisited a popular
graph-based ranking algorithm and refined how node (aka sentence)
centrality is computed. We employed BERT to better capture sentence
similarity and built graphs with directed edges arguing that the
contribution of any two nodes to their respective centrality is
influenced by their relative position in a document.  Experimental
results on three news summarization datasets demonstrated the
superiority of our approach against strong baselines. In the future,
we would like to investigate whether some of the ideas introduced in
this paper can improve the performance of supervised systems as well
as sentence selection in multi-document summarization.


\section*{Acknowledgments}

The authors gratefully acknowledge the
financial support of the European Research Council (Lapata; award
number 681760). This research is based upon work supported in part by
the Office of the Director of National Intelligence (ODNI),
Intelligence Advanced Research Projects Activity (IARPA), via contract
FA8650-17-C-9118.  The views and conclusions contained herein are
those of the authors and should not be interpreted as necessarily
representing the official policies or endorsements, either expressed
or implied, of the ODNI, IARPA, or the U.S. Government. The
U.S. Government is authorized to reproduce and distribute reprints for
Governmental purposes notwithstanding any copyright annotation
therein. \\

\bibliography{acl2019}
\bibliographystyle{acl_natbib}

\appendix

\section{Appendix}
\label{sec:appendix}

\subsection{Examples of System Output}

Table \ref{table:system_output} shows examples of system output. Specifically, we show summaries produced from  \textsc{Gold}, \textsc{Lead}, \textsc{TextRank} and \textsc{PacSum} for test documents in NYT, CNN/Daily Mail and TTNews. \textsc{Gold} is the gold summary associated with each document; \textsc{Lead} extracts the first document sentences; TextRank \cite{W04-3252} adopts PageRank \cite{ilprints361} to compute node centrality recursively based on a Markov chain model; \textsc{PacSum} is position augmented centrality based summarization approach introduced in this paper. 

\begin{table*}[t]
\begin{small}
\begin{tabular}{m{0.1cm}m{5.4cm}m{4.5cm}m{4.2cm}}

\multicolumn{1}{c}{} &  \multicolumn{1}{c}{{\bf NYT}} & \multicolumn{1}{c}{{\bf CNN+DM}} & \multicolumn{1}{c}{{\bf TTNews}} \\
\thickhline
\rotatebox{90}{\textsc{Gold}}  & Marine Corps says that V-22 Osprey, hybrid aircraft with troubled past, will be sent to Iraq in September, where it will see combat for first time.  \vspace{0.12cm} \hfill\break 
 The Pentagon has placed so many restrictions on how it can be used in combat that plane -- which is able to drop troops into battle like helicopter and then speed away like airplane -- could have difficulty fulfilling marines longstanding mission for it.  \vspace{0.12cm} \hfill\break
 Limitations on v-22, which cost \$80 million apiece, mean it can not evade enemy fire with same maneuvers and sharp turns used by helicopter pilots. 
 
& "We're all equal, and we all deserve the same fair trial." says one juror.  \vspace{0.12cm} \hfill\break
The months-long murder trial of Aaron Hernandez brought jurors together.  \vspace{0.12cm} \hfill\break
Foreperson: "It's been an incredibly emotional toll on all of us." 

&  \begin{CJK*}{UTF8}{gbsn} 皇马今夏清洗名单曝光，三小将租借外出，科恩特朗、伊利亚拉门迪将被永久送出伯纳乌球场.
 (Real Madrid's cleaning list was exposed this summer, and the three players will be rented out. Coentrao and Illarramendi will permanently leave the Bernabeu Stadium.
) \end{CJK*} \\ \hline

\rotatebox{90}{\textsc{TextRank}}  
& The Pentagon has placed so many restrictions on how it can be used in combat that the plane -- which is able to drop troops into battle like a helicopter and then speed away from danger like an airplane -- could have difficulty fulfilling the marines ' longstanding mission for it.  \vspace{0.12cm} \hfill\break 
Because of these problems, Mr. Coyle, the former pentagon weapons tester, predicted the marines will use the v-22 to ferry troops from one relatively safe spot to another, like a flying truck.
\vspace{0.12cm} \hfill\break 
In December 2000, four more marines, including the program's most experienced pilot, were killed in a crash caused by a burst hydraulic line and software problems.

& A day earlier, Strachan, the jury foreperson, announced the first-degree murder conviction in the 2013 shooting death of Hernandez's onetime friend Odin Lloyd. \vspace{0.12cm} \hfill\break 
Before the trial, at least one juror -- Rosalie Oliver -- had n't heard of the 25-year-old defendant who has now gone from a \$ 40 million pro-football contract to a term of life without parole in a maximum-security prison. \vspace{0.12cm} \hfill\break 
Rosalie Oliver -- the juror who had n't heard of Hernandez before the trial -- said that, for her, the first shot was enough.

& \begin{CJK*}{UTF8}{gbsn}
2个赛季前，皇马花费3500万欧元引进了伊利亚拉门迪，巴斯克人在安切洛蒂手下就知道，他在皇马得不到好机会，现在主教练换成了贝尼特斯，情况也没有变化。(Two seasons ago, Real Madrid spent 35 million euros to introduce Illarramendi. The Basques knew under Ancelotti that he could not get a good chance in Real Madrid. Now the head coach has changed to Benitez. The situation has not changed.) 
\end{CJK*} \\ \hline

\rotatebox{90}{\textsc{Lead}}
& the Marine Corps said yesterday that the V-22 Osprey, a hybrid aircraft with a troubled past, will be sent to Iraq this September, where it will see combat for the first time. \vspace{0.12cm} \hfill\break
But because of a checkered safety record in test flights, the v-22 will be kept on a short leash. \vspace{0.12cm} \hfill\break
The Pentagon has placed so many restrictions on how it can be used in combat that the plane -- which is able to drop troops into battle like a helicopter and then speed away from danger like an airplane -- could have difficulty fulfilling the marines ' longstanding mission for it. 

& (CNN) After deliberating for more than 35 hours over parts of seven days, listening intently to the testimony of more than 130 witnesses and reviewing more than 400 pieces of evidence, the teary-eyed men and women of the jury exchanged embraces. \vspace{0.12cm} \hfill\break
Since late January, their work in the Massachusetts murder trial of former NFL star Aaron Hernandez had consumed their lives. \vspace{0.12cm} \hfill\break
It was nothing like ``Law \& Order.''

& \begin{CJK*}{UTF8}{gbsn} 新浪体育显示图片厄德高新赛季可能会被皇马外租，皇马主席弗罗伦蒂诺已经获得了贝尼特斯制定的“清洗黑名单”。(Sina Sports shows that Ödegaard this season may be rented by Real Madrid, Real Madrid President Florentino has obtained the "cleansing blacklist" developed by Benitez.)
\end{CJK*} \\ \hline

\rotatebox{90}{\textsc{PacSUm}} 
& The Marine Corps said yesterday that the V-22 Osprey, a hybrid aircraft with a troubled past, will be sent to Iraq this September, where it will see combat for the first time. \vspace{0.12cm} \hfill\break
The Pentagon has placed so many restrictions on how it can be used in combat that the plane — which is able to drop troops into battle like a helicopter and then speed away from danger like an airplane — could have difficulty fulfilling the Marines’ longstanding mission for it. \vspace{0.12cm} \hfill\break
The limitations on the V-22, which cost \$80 million apiece, mean it cannot evade enemy fire with the same maneuvers and sharp turns used by helicopter pilots. 

& (CNN) After deliberating for more than 35 hours over parts of seven days, listening intently to the testimony of more than 130 witnesses and reviewing more than 400 pieces of evidence, the teary-eyed men and women of the jury exchanged embraces.  \vspace{0.12cm} \hfill\break
Since late January, their work in the Massachusetts murder trial of former NFL star Aaron Hernandez had consumed their lives. \vspace{0.12cm} \hfill\break
"It 's been an incredibly emotional toll on all of us." Lesa Strachan told CNN 's Anderson Cooper Thursday in the first nationally televised interview with members of the jury. 

& \begin{CJK*}{UTF8}{gbsn}
厄德高、卢卡斯-席尔瓦和阿森西奥将被租借外出，而科恩特朗和伊利亚拉门迪，则将被永久送出伯纳乌球场。(Ödegaard, Lucas Silva and Asencio will be rented out, while Coentrao and Illarramendi will permanently leave the Bernabeu Stadium.)
\end{CJK*} \\ \hline

\thickhline 

\end{tabular}
\end{small}
\caption{Example gold summaries and system output for NYT, CNN/Daily Mail and TTNews documents. } \label{table:system_output}

\vspace{-0.2in}
\end{table*}

\end{document}